\documentclass[lettersize,journal]{IEEEtran}
\usepackage{amsmath,amsfonts}
\usepackage{algorithmic}
\usepackage{algorithm}
\usepackage{array}
\usepackage[caption=false,font=normalsize,labelfont=sf,textfont=sf]{subfig}
\usepackage{textcomp}
\usepackage{stfloats}
\usepackage{url}
\usepackage{verbatim}
\usepackage{graphicx}
\usepackage{cite}
\usepackage{soul}
\usepackage{xcolor}

\usepackage[switch]{lineno}
\usepackage{etoolbox}

\begin{document}

\title{Design and Characterization of a Dual-DOF Soft Shoulder Exosuit with Volume-Optimized Pneumatic Actuator}

\author{Rui Chen$^{1*}$,  Domenico Chiaradia$^{1}$, Daniele Leonardis$^{1}$, Antonio Frisoli$^{1}$

\thanks{This project was funded by MSCA-DN / Project 101073374 - ReWIRE. Views and opinions expressed are however those of the author(s) only and do not necessarily reflect those of the European Union or the European Research Executive Agency (REA). Neither the European Union nor the granting authority can be held responsible for them.}%

\thanks{$^{1}$ The authors are with the Institute of Mechanical Intelligence and Department of Excellence in Robotics \& AI, Scuola Superiore Sant'Anna (SSSA), 56127 Pisa, Italy (corresponding email: rui.chen@santannapisa.it)}%
}

\maketitle
\begin{abstract}

Portable pneumatic systems for 2 degree-of-freedom (DOF) soft shoulder exosuits remain underexplored, and face fundamental trade-offs between torque output and dynamic response that are further compounded by the need for multiple actuators to support complex shoulder movement. This work addresses these constraints through a volume-optimized spindle-shaped angled actuator (SSAA) geometry: by reducing actuator volume by 35.7\% (357~mL vs.\ 555~mL), the SSAA maintains 94.2\% of output torque while achieving 35.2\% faster dynamic response compared to uniform cylindrical designs. Building on the SSAA, we develop a curved abduction actuator (CAA) based on the SSAA geometry and a horizontal adduction actuator (HAA) based on the pouch motor principle, integrating both into a dual-DOF textile-based shoulder exosuit (390~g). The exosuit delivers multi-modal assistance spanning shoulder abduction, flexion, and horizontal adduction, depending on the actuation.

User studies with 10 healthy participants reveal that the exosuit substantially reduces electromyographic (EMG) activity across both shoulder abduction and flexion tasks. For abduction with HAA only, the exosuit achieved up to 59\% muscle activity reduction across seven muscles. For flexion, both the single-actuator configuration (HAA only) and the dual-actuator configuration (HAA\,+\,CAA) reduced EMG activity by up to 63.7\% compared to no assistance. However, the incremental benefit of adding the CAA to existing HAA support was limited in healthy users during flexion, with statistically significant additional reductions observed only in pectoralis major. These experimental findings characterize actuator contributions in healthy users and provide design guidance for multi-DOF exosuit systems. 

\end{abstract}


\section{Introduction}
\IEEEPARstart{I}{mpairments} in shoulder motor functions significantly limit upper-limb manipulation and the ability to perform activities of daily living (ADLs)~\cite{kibler2001shoulder_rehabilitation_review}. Such limitations can arise at any age due to a wide range of conditions, including stroke~\cite{belagaje2017stroke-Rehabiliation,galiana2012Robot_stoke_shoulder}, spinal cord injury (SCI)~\cite{simpson2012SCI_life, duran2001SCI_effects}, and amyotrophic lateral sclerosis (ALS)~\cite{majmudar2014rehabilitation_shouldr_ALS}. Reduced shoulder range of motion (ROM) not only diminishes autonomy and social participation but also increases healthcare costs and caregiver burden. Addressing these challenges motivates the development of effective interventions to restore shoulder mobility and support functional independence~\cite{escamilla2009shoulder_exercise_review2}.

Given this background, assistive robotics has demonstrated its potential in rehabilitation over the past decades~\cite{chang2013robot_rehabilitation,gassert2018rehabilitation_robotic_review}. On the basis of the actuation mechanisms, shoulder assistive robots can be broadly categorized into rigid exoskeletons and soft exosuits~\cite{majidi2021review_shoudler_soft_rigid}. Rigid exoskeletons employ rigid links and motorized joints to provide precise torque control and position tracking. Representative systems such as Alex~\cite{pirondini2016Alex}, ARMin~\cite{nef2009ARMIN-exoskeleton}, T-WREX~\cite{housman2009randomized_T-WREX}, MIT-Manus~\cite{krebs2004rehabilitation_MITManus}, and CADEN-7~\cite{perry2007upper_CADEN-7} have demonstrated clinical efficacy in upper-limb rehabilitation. These systems offer advantages including high output torque, accurate trajectory control, and comprehensive workspace coverage~\cite{bardi2022upper_limb_review}. However, several limitations restrict their potential adoption for home-based rehabilitation: (1) rigid structures introduce significant inertia and mechanical constraints, possibly interfering with natural shoulder biomechanics~\cite{proietti2023_shoulder_exosuit_restoring_Science,sharma2024_review_soft_wearble_robotic,majidi2021review_shoudler_soft_rigid}; (2) kinematic misalignment between rigid joints and biological anatomy causes discomfort and pressure concentration at interface points; (3) bulky design and high cost limit accessibility; and (4) requirement for clinical supervision during setup and calibration makes them impractical for intensive, frequent home therapy.

To address these limitations, soft exosuits have emerged as a promising alternative paradigm. Soft exosuits employ compliant materials and flexible actuation to assist shoulder movement while maintaining inherent safety and kinematic transparency~\cite{Xiloyannis2022Soft_robotic_suit_review}. Unlike rigid systems, soft exosuits work in parallel with biological muscles rather than constraining joint kinematics, enabling more natural movement patterns~\cite{proietti2023_shoulder_exosuit_restoring_Science}. The lightweight design of wearable components and lower cost facilitate portable deployment for home-based therapy~\cite{Campioni2025_Preliminary_clinical_Evaluation}. Moreover, soft exosuits can be easily donned and doffed independently, supporting high-frequency training~\cite{O_2020_exosuit_therapist_fatigue}.

Among soft wearable robotic systems, shoulder exosuits that provide assistance to the proximal shoulder joint are crucial for upper-arm movement and essential for hand operation~\cite{bardi2022upper_limb_soft_wearable_devices_review}. Two main strategies exist for providing shoulder assistance: cable-driven approaches and direct pneumatic actuation~\cite{hussain2024_review_advancements_oft_wearable_robots,sharma2024_review_soft_wearble_robotic}.

Cable-driven shoulder exosuits offer notable advantages due to their lightweight, compliant, and flexible transmission~\cite{Xiloyannis2022Soft_robotic_suit_review}. By relocating actuators away from the shoulder and using cables to transmit forces, they effectively reduce inertia and mechanical constraints around the joint, improving comfort and wearability~\cite{majidi2021review_shoulder_joint}. Electrical motors enable precise position and speed control, making them the most popular solution for cable-driven actuation~\cite{georgarakis2022_cable_driven_2DOF_NMI,Samper_2020_cable_driven_2DOF,li2018_2DOF_cable_driven, Pyeon_2024_cable_driven_1DOF}. Other soft actuator solutions show potential for shoulder assistance, such as shape memory alloys~\cite{golgouneh2021design_SMA,park2025_cable_driven_SMA} and pneumatic artificial muscles~\cite{natividad2020_cable_driven-2DOF_PAMs}, due to their higher power-to-weight ratio, intrinsic compliance, and safer human–robot interaction.

Fabric-based actuators, typically made of thermoplastic polyurethane (TPU)-coated materials, offer the advantages of compliance, lightweight design, and inherent safety~\cite{Rui2025Glove,Rui2025_Thermal_haptic,Rui2025hapmorph_Haptic, Rui2025LPPAMs, li2022_soft-actuator_review,ferroni2026ShoulderExosuitOptimation}. Direct pneumatic actuators based on fabrics provide torque assistance directly to the joint~\cite{natividad2016_development_shoulder_upper}. Stacking pouch motors together to create shoulder abduction assistance results in torque that decreases almost linearly as abduction increases, while more torque is actually needed to compensate for gravity~\cite{yao2025design_pouch_stacking, Sahin_2022_fold_pouch_motor}. Bar-shaped actuators act like structural bars between the shoulder and chest~\cite{simpson2020_exomuscle_bar,ayazi2025analysis_force_sexertedshoulder_bar}; however, limited contact area results in discomfort. Pre-curved pneumatic actuators with predefined bending angles have emerged as a promising design trend, as their geometry ergonomically conforms to the anatomy of the axillary region while enabling more compact wearable integration~\cite{o2022unfolding_modeling_prebending, simpson2017exomuscle_prebending, Campioni2025_Preliminary_clinical_Evaluation}.

However, only a few exosuits provide 2 degrees of freedom (DOF) at the shoulder joint, offering not only shoulder abduction but also shoulder flexion~\cite{o2017_shoulder_Design_characterization_preliminary_testing_2DOF, arellano2019_soft_pneumatic_2DOF}. Shoulder flexion is essential for many activities of daily living (ADLs), such as eating, drinking, and using a phone~\cite{oosterwijk2018shoulder_ROM_ADL, gates2016ROM_ADLs}, and active flexion assistance proves important for rehabilitation applications as well~\cite{martin2024SCI_data,simpson2012SCI_life,belagaje2017stroke-Rehabiliation}.

While dual-DOF pneumatic soft shoulder exosuits offer functional benefits, driving multiple actuators introduces critical system-level constraints for portable pneumatic systems---typically comprising small compressors with reservoirs regulated via solenoid valves~\cite{zhou2024portable_industry, Campioni2025_Preliminary_clinical_Evaluation, proietti2023_shoulder_exosuit_restoring_Science}. Each additional actuator demands greater air consumption, directly reducing the number of achievable repetitions for a given portable supply~\cite{joshi2021Pneumatic_supply_system, zhou2024portable_industry}. To ensure adequate repetitions during use, the system must therefore compromise either dynamic response for better portability, torque output by adopting smaller actuators, or accept increased weight, size, and power consumption from a more powerful air supply. Actuator optimization is thus essential to maximize torque output and dynamic response without increasing the size and weight of the portable pneumatic system.

In this work, we present a design methodology for dual-DOF soft shoulder exosuits constrained by portable pneumatic systems. The exosuit incorporates a curved abduction actuator (CAA) and a fabric horizontal adduction actuator (HAA), providing active assistance for both shoulder abduction and shoulder flexion. We propose a spindle-shaped geometry that addresses volume-dynamic response trade-offs. A user study with healthy participants characterizes the functional contributions of individual actuators.

This work makes three main contributions to the design of portable pneumatic exosuits:

\textbf{C1. Volume-optimized actuator design methodology:} 
We introduce a systematic approach to pneumatic actuator geometry optimization based on moment arm analysis, demonstrating that non-uniform cross-sections enable 35.7\% volume reduction while maintaining 94.2\% torque output in operational ranges ($\theta_a > 90^\circ$) and achieving 35.2\% faster response.

\textbf{C2. Dual-DOF exosuit system integration and validation:} 
We present design, fabrication, and benchtop characterization of a textile-based dual-DOF shoulder exosuit, including analytical models horizontal adduction actuators, with experimental validation for CAA across torque-angle characteristics, frequency response, and durability.

\textbf{C3. Empirical characterization of exosuit assistance:} 
We provide systematic EMG-based evaluation across seven muscles, demonstrating that the exosuit achieves up to 59\% muscle activity reduction during abduction tasks. For flexion, both single-actuator and dual-actuator configurations substantially reduce muscle activity (up to 63.7\%), while the incremental benefit of adding horizontal adduction actuator is limited to specific muscle groups in healthy users.

\begin{figure*}[tbhp]
\centering
\includegraphics[width=1\linewidth]{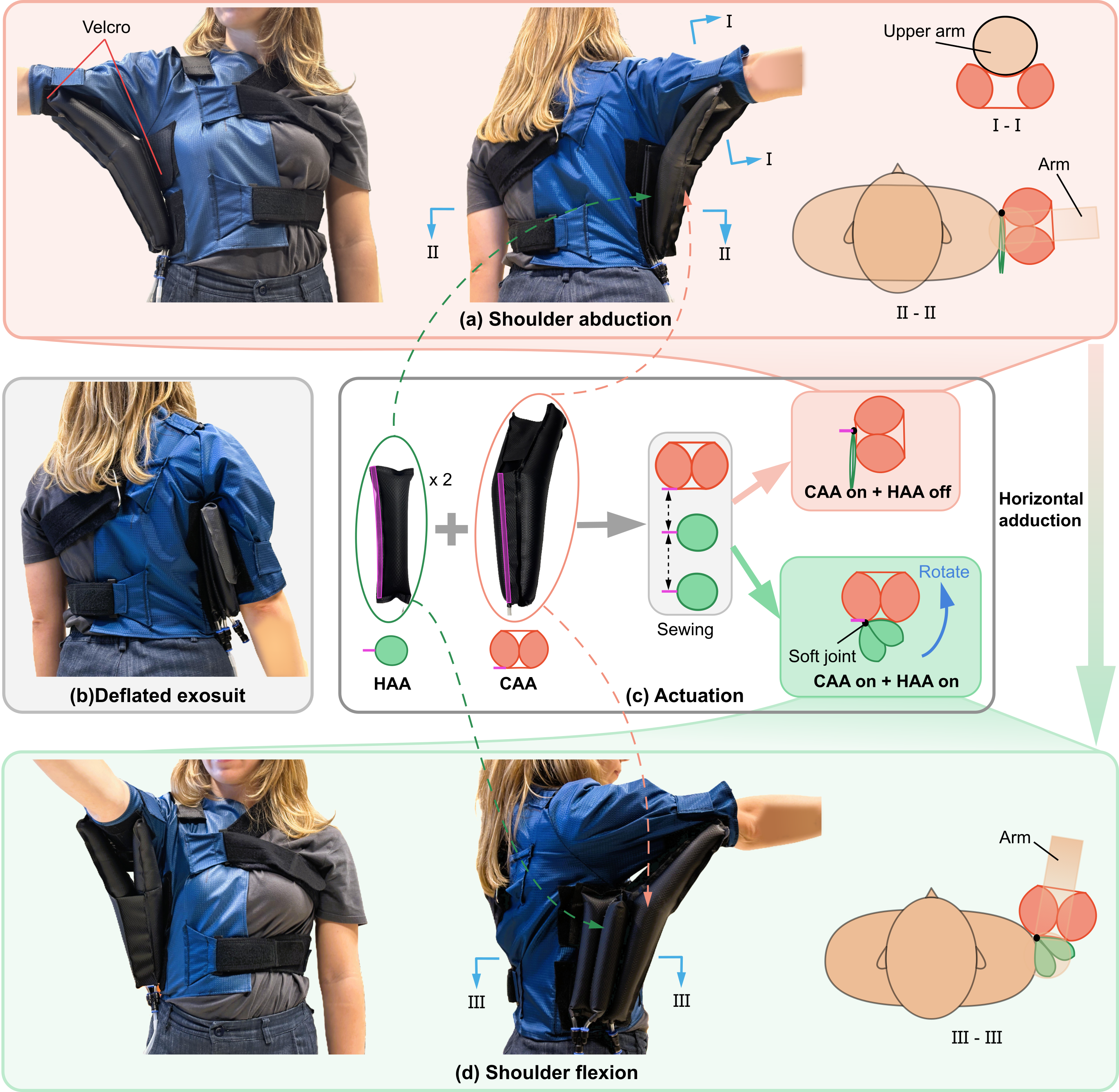}
\caption{Conceptual overview of the shoulder exosuit. (a) Shoulder abduction assistance. (b) Exosuit in deflated state. (c) Actuators of the exosuit. (d) Shoulder flexion assistance.}
\label{Conceptual}
\end{figure*}

\section{Design and Characterization of the Shoulder Exosuit}

\subsection{Exosuit Design}

The exosuit consists of a half-shirt for wearability, the CAA for abduction, and the HAA for horizontal adduction (Fig.~\ref{Conceptual}c). The half-shirt features a blue fabric base and two black straps with Velcro fasteners for size adjustment and adaptability. The actuators are sewn together on one side to the half-shirt between the arm and chest, allowing soft rotation around the sewn edge. The sewn actuators are then attached to the half-shirt via Velcro, enabling rapid actuator replacement. The wearable portion of the soft exosuit weighs 390 grams. When deflated, the actuators collapse into thin layers of fabric, minimizing space and resistance during neutral shoulder movement (Fig.~\ref{Conceptual}b).

The exosuit can support three shoulder movements depending on the actuation sequence of the two actuators: flexion, abduction, and horizontal adduction (transition from shoulder abduction to forward flexion via horizontal adduction). When only the CAA is inflated, the exosuit provides natural abduction assistance, while allowing the user to perform flexion and horizontal adduction with their own muscular activation (Fig.~\ref{Conceptual}a). When both CAA and HAA are actuated simultaneously, the exosuit provides active shoulder flexion assistance (Fig.~\ref{Conceptual}d). If the HAA is actuated after the CAA, shoulder horizontal adduction is assisted (transition from Fig.~\ref{Conceptual}a to Fig.~\ref{Conceptual}d).

\subsection{Abduction Actuator Design}

\begin{figure*}[tbhp] 
\centering 
\includegraphics[width=1\linewidth]{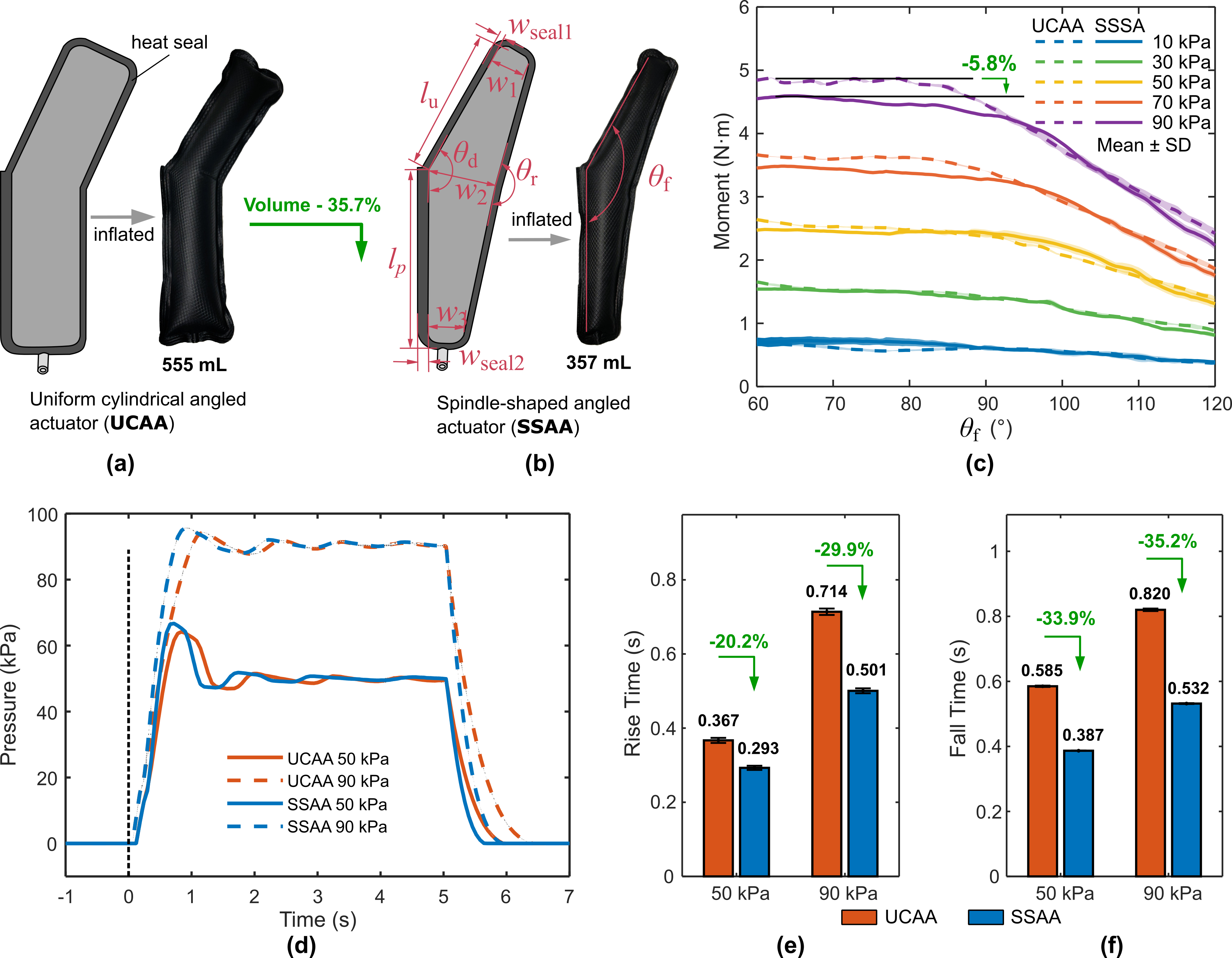} 
\caption{Comparison of uniform cylindrical angled actuator (UCAA) and spindle-shaped angled actuator (SSAA). (a) Illustration and dimensions of the two actuators. (b) Moment-angle relationship of the two actuators under different pressures. (c) Step response of the two actuators at 50 kPa and 90 kPa. (d) Rise time and (e) fall time comparison of the two actuators.}
\label{Comparsion} 
\end{figure*}

Dynamic response
in pneumatic systems is constrained by both actuator volume and air supply system capability. To ensure necessary dynamics for the exosuit, portable pneumatic devices must either increase pump and air reservoir size or decrease actuator dimensions directly, resulting in a fundamental trade-off between system weight and output performance.

Current actuator solutions for shoulder exosuits typically use cylindrical actuators with a pre-curved angle for better conformity to the anatomical shape of the axillary region~\cite{proietti2023_shoulder_exosuit_restoring_Science,Campioni2025_Preliminary_clinical_Evaluation}, which we refer to as the uniform cylindrical angled actuator (UCAA) based on its uniform cross-section, as shown in Fig.~\ref{Comparsion}(a). However, for bending actuators, generating the same torque requires smaller force as distance from the bending point increases. In pneumatic bending actuators, when the bending angle varies within a small range, the resulting bending moment is primarily determined by the contact conditions of the bending section. Only when the bending angle decreases below a certain threshold, the cross-sections at both ends begin to significantly influence the generated torque~\cite{o2022unfolding_modeling_prebending}.

Based on these mechanical principles, we propose that a bending actuator does not require uniform cross-sectional area along its length. Instead, a cross-section that gradually decreases toward both ends can accommodate the smaller force required at distal regions while effectively reducing overall actuator volume. Accordingly, we introduce a volume-optimized actuator, termed the spindle-shaped angled actuator (SSAA), characterized by cross-sectional area that progressively decreases with increasing distance from the bending section, which maintains a pre-curved angle, as illustrated in Fig.~\ref{Comparsion}(b).

The nomenclature and dimensions for UCAA and SSAA are presented in Table~\ref{tab:actuator_parameters}. This parameter design was inspired by prior work~\cite{zhou2024portable_industry, proietti2023_shoulder_exosuit_restoring_Science}, with modifications made to accommodate our system. Compared to the UCAA, the SSAA has the same middle width $w_2$ for consistent output but smaller upper width $w_1$ and bottom width $w_3$ for volume reduction. The parameters $\theta_d$ and $\theta_r$ are correspondingly adjusted so that the UCAA and SSAA achieve similar inflated external angles $\theta_f$ after inflation (151$^\circ$ for UCAA and 148$^\circ$ for SSAA). As a result, the actuator volume is 555~mL for UCAA and 357~mL for SSAA, representing a 35.7\% volume reduction (estimated from simplified 3D models based on the dimensions).

\begin{table}[tbhp]
\centering
\caption{Nomenclature and Dimensions of UCAA and SSAA Actuators}
\label{tab:actuator_parameters}
\begin{tabular}{cllcc}
\hline
\textbf{Nomenclature} & \textbf{Description} & \textbf{Unit} & \textbf{UCAA} & \textbf{SSAA} \\
\hline
$l_u$ & Upper segment length & mm & 162 & 162 \\
$l_p$ & Lower segment length & mm & 192 & 192 \\
$w_1$ & Upper width & mm & 90 & 52 \\
$w_2$ & Middle width & mm & 90 & 90 \\
$w_3$ & Bottom width & mm & 90 & 56.5 \\
$w_{\text{seal1}}$ & Sealing width & mm & 7.5 & 7.5 \\
$w_{\text{seal2}}$ & Stitching sealing width & mm & 10 & 10 \\
$\theta_d$ & Designed external angle & deg & 155 & 150 \\
$\theta_r$ & Actual internal angle & deg & 155 & 177 \\
$\theta_f$ & Inflated external angle & deg & 151 & 148 \\
\hline
\end{tabular}
\end{table}

We further compared the output moment of the UCAA and SSAA at different pressure levels (using the same experimental setup for the CAA described in the next section), and the results are shown in Fig.~\ref{Comparsion}(c). For inflated external angles $\theta_f$ between 90$^\circ$ and 120$^\circ$, the moment difference between the UCAA and SSAA is negligible across all pressure levels. For angles between 60$^\circ$ and 90$^\circ$, the difference remains negligible at lower pressures, but the SSAA exhibits a 5.8\% maximum moment reduction compared to the UCAA at 90~kPa (4.88~Nm for UCAA versus 4.60~Nm for SSAA). This can be attributed to increased crease area at lower angles $\theta_f$ due to the larger dimensions of $w_1$ and $w_3$~\cite{o2022unfolding_modeling_prebending}.

We also compared the step response of the two actuators using a customized pneumatic control system (the same one used for the user study, with details provided in the user study section). We set the pressure reference for the ITV0030 proportional pressure regulators and measured real-time pressure with embedded pressure sensors using analog signals (supply pressure from the air compressor was 130~kPa). We provided step pressure references of 50~kPa and 90~kPa to the actuators and measured the inflated pressure in real time. The experiments were conducted with three repetitions at each pressure level, and we present the mean with standard deviations (SD).

The SSAA demonstrated faster response compared to the UCAA, as shown in Fig.~\ref{Comparsion}(d-e). For rise time, the SSAA showed 20.2\% reduction at 50~kPa and 29.9\% reduction at 90~kPa. For fall time, the reductions were 33.9\% at 50~kPa and 35.2\% at 90~kPa, which closely matches the volume reduction. In conclusion, the SSAA design enables 35.7\% volume reduction and faster dynamic response up to 35.2\%, with only a 5.8\% output compromise at lower angles $\theta_a$ at high pressure.

\subsection{Abduction Actuator Fabrication and Characterization}

\begin{figure*}[tbhp] 
\centering 
\includegraphics[width=1\linewidth]{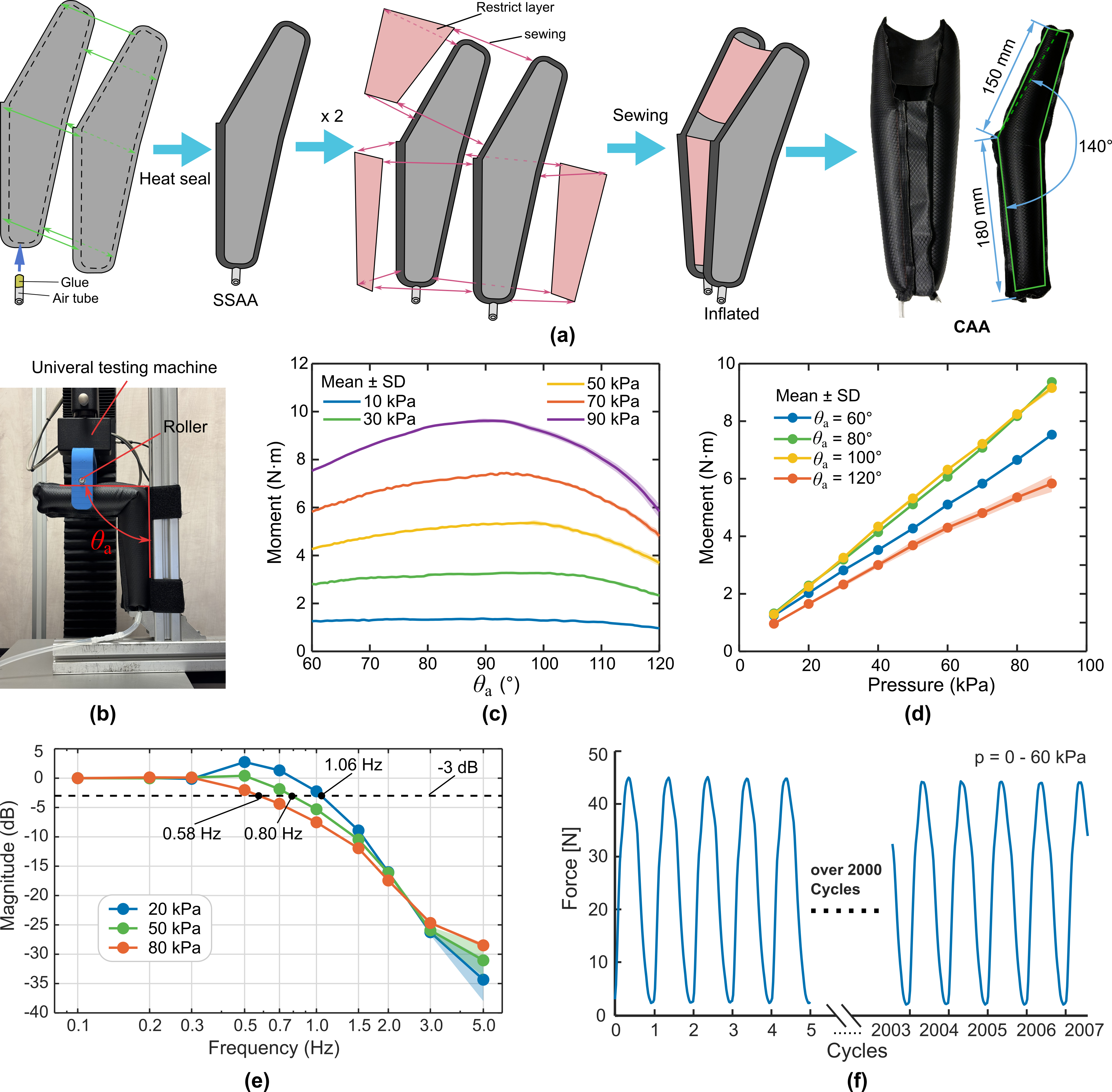} 
\caption{curved abduction actuator characterization. (a) Fabrication process of the curved abduction actuator. (b) Experimental setup for moment measurement. (c) Moment-angle relationship of the abduction actuator at various pressure levels. (d) Moment-pressure relationship of the abduction actuator. (e) Bode plot of the actuator at different pressure levels. (f) Force output stability over 2000 inflation cycles.}
\label{Abduction-Characterization} 
\end{figure*}

For shoulder abduction assistance, it has been observed that Y-shaped actuators can provide better anchoring and more balanced arm support compared to simple cylindrical shapes~\cite{sahin2024Shoulder_anchoring_configurations, li2022development_Y-shape, simpson2017exomuscle_prebending, Campioni2025_Preliminary_clinical_Evaluation, proietti2023_shoulder_exosuit_restoring_Science}. To adapt the SSAA into a Y-shaped actuator for abduction, we developed the curved abduction actuator (CAA), and its fabrication process is shown in Fig.~\ref{Abduction-Characterization}(a). Two layers of TPU-coated fabric were laser-cut and heat-sealed to form the SSAA. Three trapezoidal constraint layers were then introduced to guide the inflation of the two SSAAs. The constraint layers are narrower near the bottom (waist) and progressively wider toward the top (arms), shaping the CAA into a Y-shaped configuration. After inflation, the CAA achieves a CAA pre-curved angle $\theta_a$ of 140$^\circ$, with the concave portion of the Y-shape designed for arm contact. The upper contact length for the arm is 150~mm and the lower contact length for the axillary region is 180~mm. All actuators in this work are made of 210D diamond ripstop TPU-coated fabric (Adventure Expert, Slovenia); different colors were used only for distinction and clarity.

We first characterized the CAA moment output at different bending angles $\theta_a$ and pressure levels. The experimental setup is based on a universal testing machine (Alluris FMT-313, Germany) as shown in Fig.~\ref{Abduction-Characterization}(b), similar to related work~\cite{Campioni2025_Preliminary_clinical_Evaluation}. We mounted the CAA to an aluminum frame with fabric and Velcro, maintaining a 10~cm distance between the frame and the moving center. A 3D-printed part with a roller was fixed to the universal testing machine to minimize friction between the actuator and moving parts. During testing, the roller moved from top to bottom at 5~mm/s at pressures of 10, 20, ..., 90~kPa. Force and roller position were recorded at 100~Hz, and moment output was calculated. For each pressure level, three measurement repetitions were performed, and the mean and SD are reported.

The results are shown in Fig.~\ref{Abduction-Characterization}(c-d). Moment output is more stable with changes in bending angle $\theta_a$ at 10~kPa. As pressure increases, the moment first increases with angle, reaching a maximum value around 90$^\circ$, then decreases with further angle increases. For a given bending angle, output is approximately linear with actuation pressure. Compared to a single SSAA actuator, moment decreases as angle decreases, likely due to constraint layers limiting volume change at lower angles. The maximum moment achieved is 9.7~Nm around 90$^\circ$ at 90~kPa, a comparable result to prior work~\cite{proietti2023_shoulder_exosuit_restoring_Science,Campioni2025_Preliminary_clinical_Evaluation}.

Fig.~\ref{Abduction-Characterization}(e) shows the magnitude Bode plots of the actuator measured at different input pressures of 20~kPa, 50~kPa, and 80~kPa. For each pressure condition, frequency response was measured three times, and the mean magnitude with variance is reported. At low frequencies ($<$0.3~Hz), magnitude responses remain close to 0~dB for all pressures with small variance. The -3~dB cutoff frequencies are 1.06~Hz, 0.8~Hz, and 0.58~Hz for 20~kPa, 50~kPa, and 80~kPa, respectively. Cutoff frequency decreases with increasing input pressure, demonstrating reduced effective bandwidth at higher pressures due to longer dynamic response times as discussed previously.

At higher frequencies, magnitude responses decay rapidly for all pressure levels and tend to converge, suggesting that high-frequency behavior is dominated by system dynamics rather than operating pressure. Overall, small variance across repeated measurements confirms the consistency and reliability of measured frequency responses.

To evaluate actuator durability, extensive inflation–deflation cycling experiments were conducted, as shown in Fig.~\ref{Abduction-Characterization}(f). Internal pressure was repeatedly varied between 0 and 60~kPa, and output force was measured throughout the cycling process. In total, more than 2000 inflation cycles were performed, consisting of two independent tests with 1010 cycles each. No noticeable degradation in output force was observed after cycling experiments. The force–pressure relationship remained consistent before and after cycling tests, indicating stable mechanical performance of the actuator under repeated pneumatic loading.

\subsection{Horizontal Adduction Actuator Fabrication and Characterization}

\begin{figure*}[tbhp] 
\centering 
\includegraphics[width=1\linewidth]{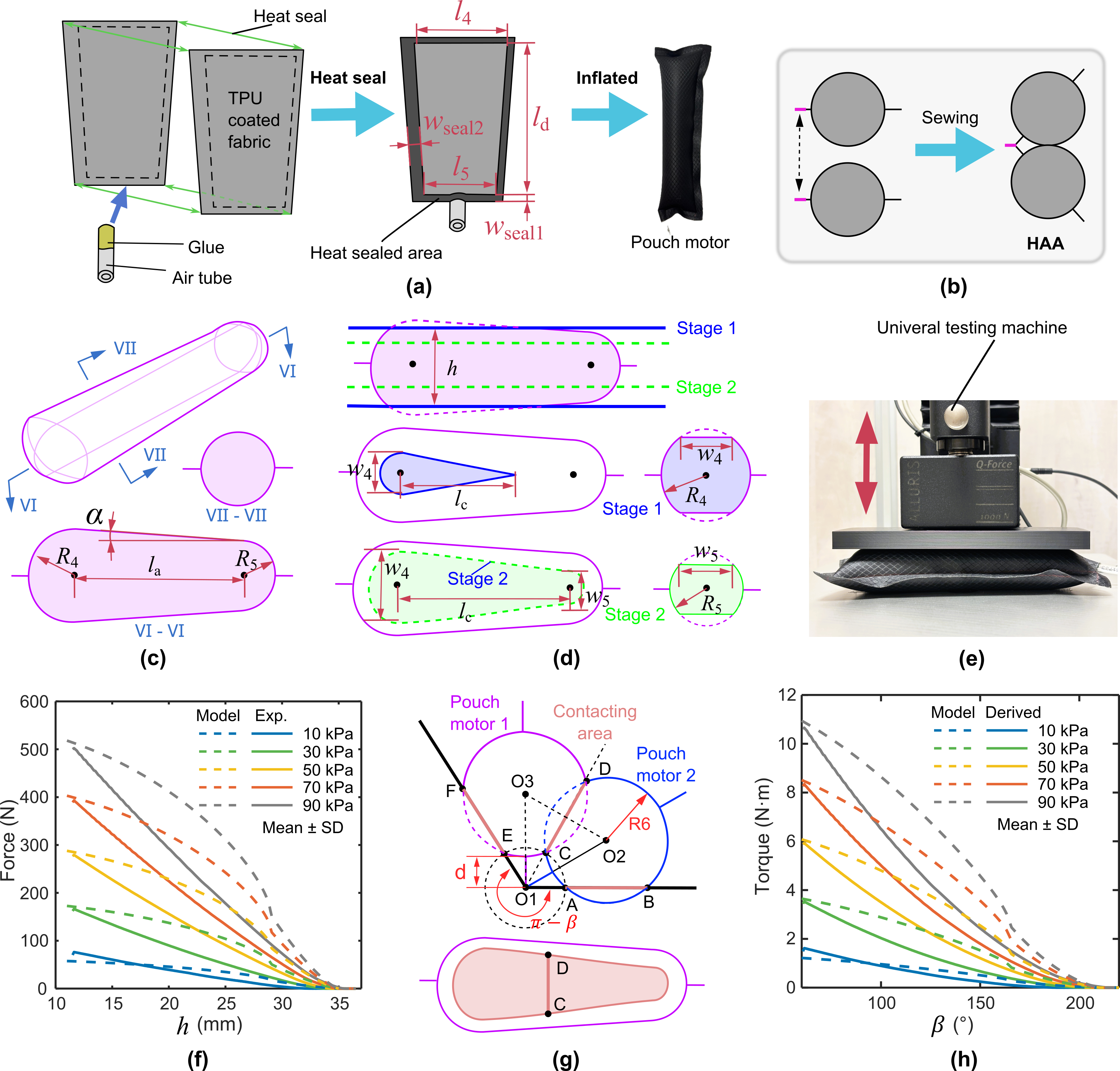} 
\caption{Fabric horizontal adduction actuator characterization. (a) Fabrication and dimensions of a pouch motor. (b) Sewing two pouch motors together to create the horizontal adduction actuator. (c) Parameters of the pouch motor. (d) Experimental setup for pouch motor characterization. (e) Parameters for pouch motor modeling. (f) Validation of the force--height model against experimental measurements at various pressure levels. (g) Geometric parameters for torque--angle modeling of the horizontal adduction actuator. (h) Comparison of modeled and experimentally derived torque as a function of horizontal adduction angle.}
\label{Flexion-Characterization} 
\end{figure*}

In this work, stacked pouch motors were adopted to actuate shoulder horizontal adduction. The pouch motors were designed with a trapezoidal profile complementary to the CAA geometry, with upper length $l_4 = 118$~mm, lower length $l_5 = 88$~mm, and height $l_d = 180$~mm, as shown in Fig.~\ref{Flexion-Characterization}(a). Two pouch motors were sewn together along one side to form the horizontal adduction actuator (HAA), enabling a larger range of angular rotation, as illustrated in Fig.~\ref{Flexion-Characterization}(b).

To characterize the torque output of the HAA as a function of horizontal adduction angle, we developed a simplified analytical model for the pouch motor. Given that force is more directly accessible experimentally than torque, the model was first formulated and validated in the force--height domain, then extended to the torque--angle relationship through the geometric coupling between actuator deformation and joint kinematics. The relevant geometric parameters are illustrated in Fig.~\ref{Flexion-Characterization}(c-d). The pouch motor is characterized by upper radius $R_4 = l_4 / \pi$ and lower radius $R_5 = l_5 / \pi$ at the proximal and distal ends, respectively, with the distance between the two contact ends $l_a = l_d - 2 w_{\text{seal1}} - R_4 - R_5$. The taper angle is defined as $\alpha = \arctan(|R_4 - R_5|/l_a)$.

When the actuator is compressed to height $h$, the contact widths at both ends are governed by:
\begin{equation}
\begin{cases}
w_{4} = 2\sqrt{R_4^2 - h^2/4} \\[5pt]
w_{5} = \begin{cases}
2\sqrt{R_5^2 - h^2/4} & \text{if } h \leq 2R_5 \\
0 & \text{otherwise}
\end{cases} \\[5pt]
l_c = \dfrac{|R_4 - h/2|}{\tan(\alpha)}
\end{cases}
\label{eq:geometric_constraints}
\end{equation}
where $w_4$ and $w_5$ are the upper and lower contact widths, respectively, and $l_c$ is the intermediate contact length. The effective contact area $A$ exhibits piecewise behavior depending on whether $l_c$ exceeds the full actuator span $l_a$:
\begin{equation}
A = 
\begin{cases}
A_1 + A_2 & \text{if } l_c \leq l_a \text{ (short contact)} \\[5pt]
A_1 + A_3 + A_4 & \text{if } l_c > l_a \text{ (extended contact)}
\end{cases}
\label{eq:contact_area}
\end{equation}
where the area components are:
\begin{equation}
\begin{cases}
A_1 = \dfrac{\pi w_{4}^2}{4} & \text{(proximal cap)} \\[8pt]
A_2 = \dfrac{l_c w_{4}}{2} & \text{(triangular section)} \\[8pt]
A_3 = \dfrac{\pi w_{5}^2}{4} & \text{(distal cap)} \\[8pt]
A_4 = \dfrac{(w_{4} + w_{5})l_a}{2} & \text{(trapezoidal section)}
\end{cases}
\label{eq:area_components}
\end{equation}

The contact force is then computed as:
\begin{equation}
F(h, P) = 0.001 \times P \times A(h)
\label{eq:force_height}
\end{equation}
where $P$ is the internal pressure in kPa, $F$ is the output force in N, and the factor $0.001$ converts kPa$\cdot$mm$^2$ to N.

The model was validated against experimental measurements using the setup shown in Fig.~\ref{Flexion-Characterization}(e). A universal testing machine compressed the actuator at 200~mm/min across a range of pressure levels, recording force and displacement simultaneously. Three repetitions were performed at each pressure level and results are reported as mean $\pm$ SD. As shown in Fig.~\ref{Flexion-Characterization}(f), at $h = 12$~mm the reaction force reaches approximately 500~N at 90~kPa. The model captures the overall experimental trends, though it yields slightly higher predictions at larger compression depths; this discrepancy arises from the two-stage contact assumption, which produces a convex force--height profile whereas the experimental data exhibit a more linear relationship.

Building on the validated force--height model, the torque output is predicted as a function of bending angle $\beta$, representing the actuator operating condition within the exosuit, as shown in Fig.~\ref{Flexion-Characterization}(g). In this configuration, both the compression height and the moment arm vary simultaneously with $\beta$.

The geometry is parameterized by $l_{O_1O_2} = R_5 + d$, the distance from the pouch motor center to the sewing center, where $d = 7.5$~mm is the inter-center offset introduced by sewing. The compression height and moment arm are related to bending angle by:
\begin{equation}
\begin{cases}
h(\beta) = 2\, l_{O_1O_2} \sin\!\left(\dfrac{\beta}{4}\right) \\[8pt]
l_{\text{arm}}(\beta) = l_{O_1O_2} \cos\!\left(\dfrac{\beta}{4}\right)
\end{cases}
\label{eq:kinematics}
\end{equation}
where $l_{\text{arm}}$ is the perpendicular distance from the line of force action to the rotation center, reaching its maximum at $\beta = 0^{\circ}$ and decreasing monotonically as $\beta$ increases.

Substituting $h(\beta)$ into Eq.~\eqref{eq:force_height} yields the angle-dependent force:
\begin{equation}
F(\beta, P) = F(h(\beta), P)
\label{eq:force_angle}
\end{equation}
and the output torque follows as:
\begin{equation}
M(\beta, P) = F(\beta, P) \times l_{\text{arm}}(\beta) \times 10^{-3}
\label{eq:torque}
\end{equation}
where the factor $10^{-3}$ converts N$\cdot$mm to N$\cdot$m. The piecewise nature of Eq.~\eqref{eq:force_height} propagates into the torque--angle relationship, producing distinct regimes corresponding to transitions between the short and extended contact configurations.

The modeled and experimentally derived torque across the horizontal adduction range are compared in Fig.~\ref{Flexion-Characterization}(h). Both results consistently show torque increasing with decreasing $\beta$ and with increasing pressure. At 90~kPa, the model predicts a maximum torque of 10.93~Nm at $\beta = 60^{\circ}$, in close agreement with the experimentally derived value of 10.67~Nm under the same condition. Torque decreases to zero at $\beta = 210^{\circ}$ in both cases. At $\beta = 120^{\circ}$, the HAA delivers a derived torque of approximately 4.64~Nm at 90~kPa, confirming adequate assistance capacity across the functional range of horizontal adduction.

\section{User Study}

\subsection{Experimental Protocol and Setup}

\begin{figure*}[tbhp]
\centering
\includegraphics[width=1\linewidth]{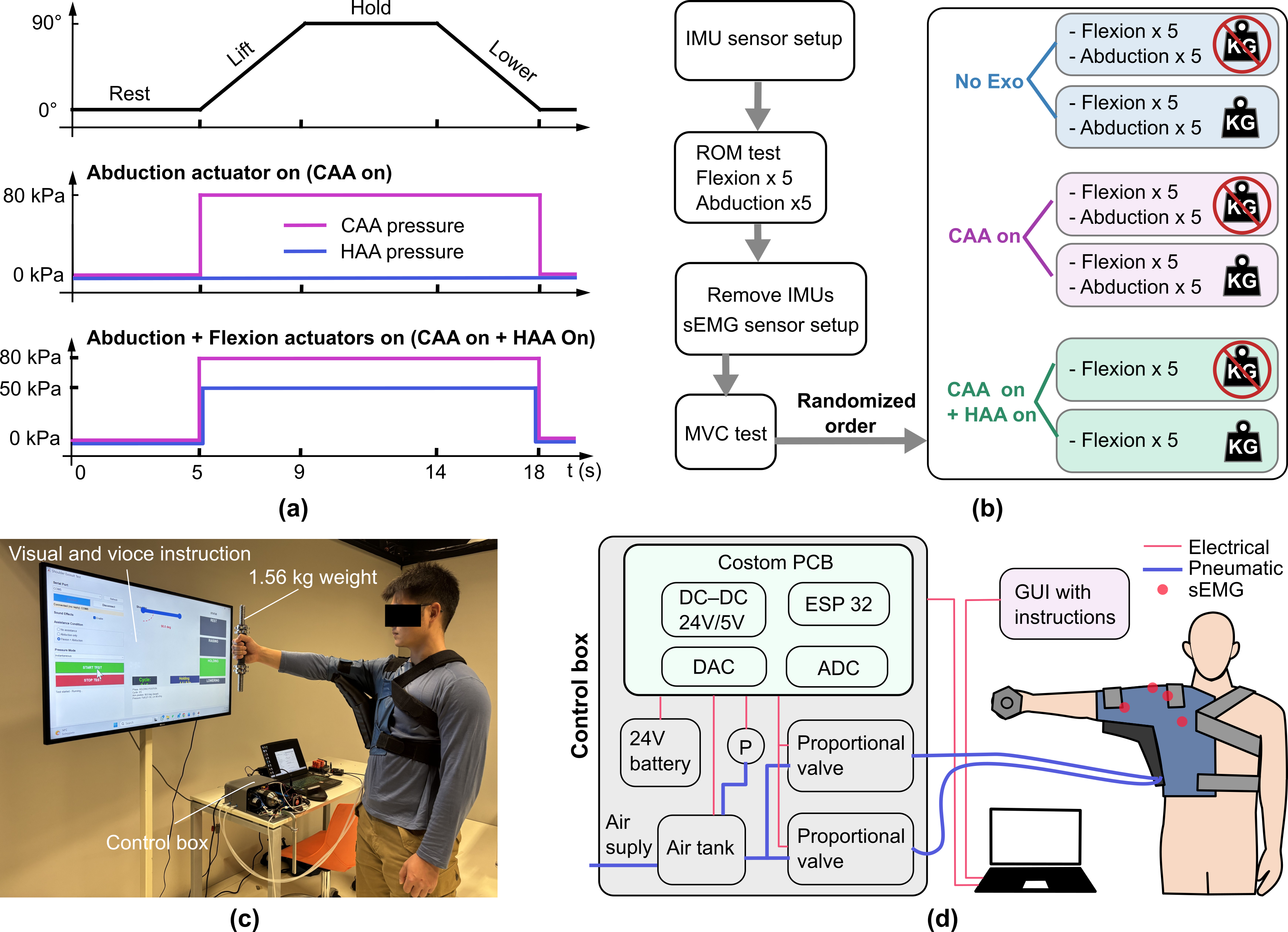}
\caption{Subject testing protocol. (a) Setup for flexion experiments. (b) Communication between system and subjects. (c) Movement and actuator pressure conditions. (d) Procedure of the subject test.}
\label{Subject-setup}
\end{figure*}

To evaluate the performance of the designed exosuit, we conducted a preliminary test with ten healthy subjects (5 M, 5 F, 27.1 ± 3.0 years old, 166.9 ± 9.6~cm height, 65.4 ± 13.9~kg weight), as shown in Table~\ref{Subjects}. All subjects reported no previous arm injury or disease, and all were right-hand dominant. The protocol was approved by the Scuola Superiore Sant'Anna Ethics Committee (No. 40/2025), and all subjects signed informed consent.

Although the ten subjects had a wide range of heights (155 to 183~cm) and weights (50 to 85~kg), one shoulder exosuit was used for all subjects, demonstrating adaptability. Subjects conducted two groups of experiments: measuring range of motion (ROM) for shoulder flexion and abduction, and measuring muscle activity during dynamic flexion and abduction tasks.

\begin{table}[htbp]
\caption{Participant Demographics}
\label{Subjects}
\centering
\begin{tabular}{ccccc}
\hline
\textbf{ID} & \textbf{Gender} & \textbf{Age (years)} & \textbf{Height (cm)} & \textbf{Weight (kg)} \\
\hline
1 & M & 34 & 178 & 74.0 \\
2 & M & 26 & 176 & 82.5 \\
3 & F & 30 & 159 & 50.0 \\
4 & F & 30 & 160 & 50.0 \\
5 & F & 27 & 156 & 58.0 \\
6 & M & 28 & 170 & 71.0 \\
7 & F & 26 & 155 & 50.0 \\
8 & M & 25 & 172 & 75.0 \\
9 & F & 24 & 156 & 49.0 \\
10 & M & 31 & 183 & 85.0 \\
\hline
\multicolumn{2}{c}{\textbf{Mean ± SD}} & 27.1 ± 3.0 & 166.9 ± 9.6 & 65.4 ± 13.9 \\
\hline
\end{tabular}
\end{table}

The shoulder ROM test was conducted to evaluate the impact of the exoskeleton on shoulder mobility. Subjects were instructed to perform maximum shoulder flexion and abduction under two conditions: without the exoskeleton and with the exoskeleton (actuators deactivated). Subjects first donned a body movement tracking system (Xsens Awinda), then performed flexion and abduction five times under each condition. Maximum angles from the five repetitions were recorded using MVN Analyze software (Xsens) as maximum ROM~\cite{norkin2016_measurement_of_joint}.

For dynamic evaluation of the exoskeleton, two groups of experiments were conducted: shoulder abduction and shoulder flexion. For shoulder abduction, two exoskeleton conditions were compared: (i) no exosuit and (ii) abduction actuator activated only (CAA on). Muscle activity was measured under two loading conditions: without external load and with a 1.56~kg weight attached to the participant's hand, resulting in four experimental conditions for shoulder abduction.

For shoulder flexion, three exoskeleton conditions were evaluated: (i) no exosuit, (ii) abduction actuator activated only (CAA on), and (iii) both abduction and horizontal adduction actuators activated simultaneously (CAA and HAA on). Each exoskeleton condition was tested under both unloaded and loaded conditions (1.56~kg), resulting in six experimental conditions for shoulder flexion.

Overall, for each subject, five exoskeleton conditions combined with two loading conditions were tested, yielding ten total experimental conditions. For each trial, participants were instructed to keep their arm naturally hanging (with or without holding the weight) for 5~s, then raise the arm to 90$^\circ$ within 4~s in either flexion or abduction, hold at 90$^\circ$ for 5~s, and subsequently lower back to the resting position within 4~s (guided by a GUI explained later). This sequence constituted one repetition, as illustrated in Fig.~\ref{Subject-setup}(a).

Under different exoskeleton conditions, different actuator pressures were applied (Fig.~\ref{Subject-setup}(a)). In the no-exosuit condition, participants did not wear the exoskeleton. In the CAA-only condition, 80~kPa pressure was supplied to the CAA. When both actuators were activated, pressures of 80~kPa and 50~kPa were applied to the CAA and HAA, respectively. Actuation pressure was applied at the onset of arm elevation and released to 0~kPa at the onset of arm lowering.

To improve measurement reliability, each subject performed five repetitions for each experimental condition. A rest period of 2–3~min was provided between consecutive trials to prevent muscle fatigue. To minimize the influence of experimental order, the sequence of the five exoskeleton conditions was fully randomized across subjects (Fig.~\ref{Subject-setup}(b)).

\subsection{Control Implementation and Actuation System}

\begin{figure*}[tbhp]
\centering
\includegraphics[width=1\linewidth]{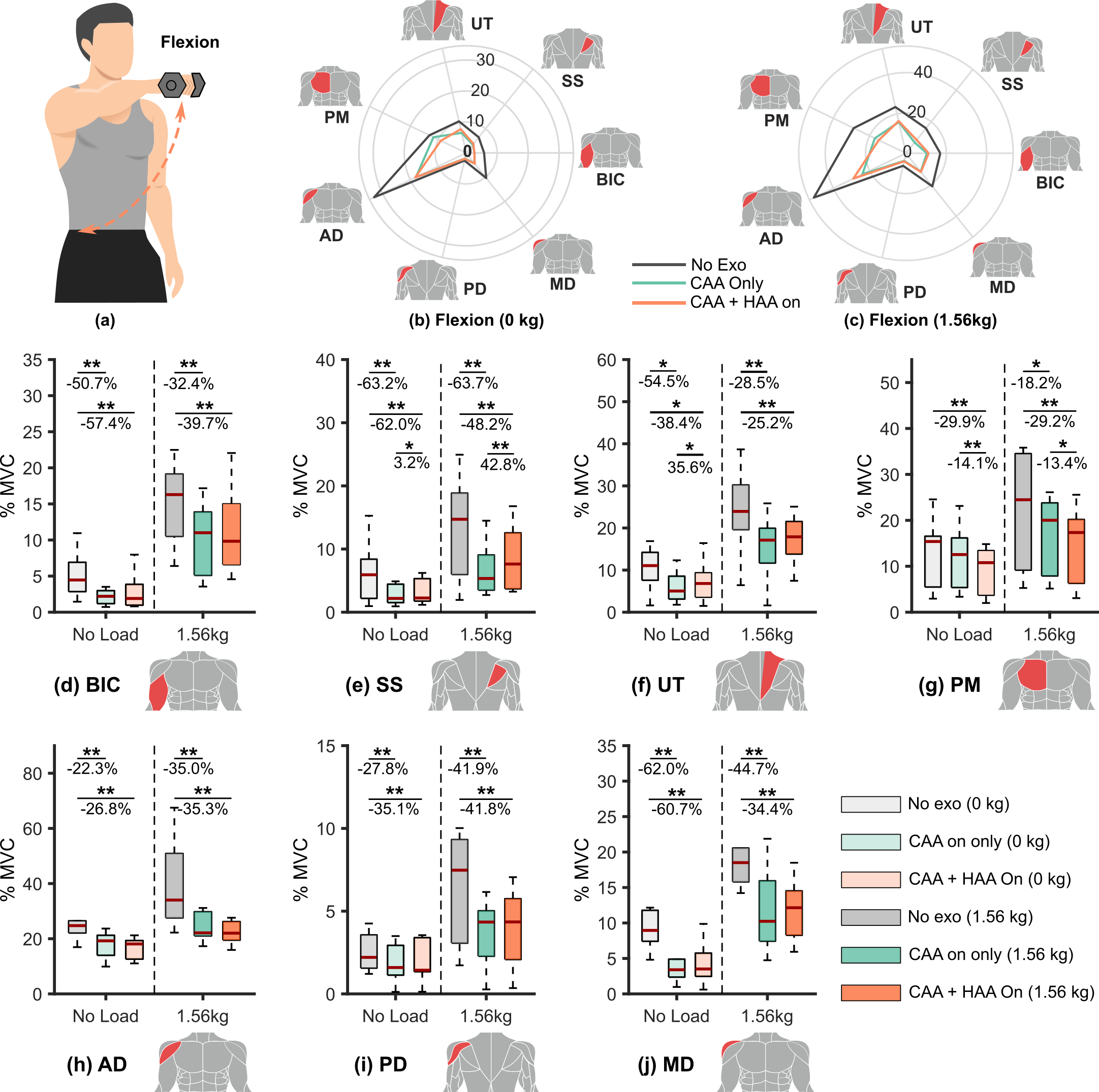}
\caption{Muscle activity comparison during shoulder flexion movements under different exosuit and weight conditions. (a) Flexion movement with weight. (b-c) Polar plot of muscle activation patterns during shoulder flexion without load and with load. (d-j) Statistical results for each muscle during shoulder flexion movements under different exosuit and weight conditions. Statistical analysis: Wilcoxon signed-rank tests with Bonferroni correction (**: $p < 0.0167$; *: $0.0167 \leq p < 0.05$).}
\label{Subject-flexion}
\end{figure*}

The control system used in experiments is illustrated in Fig.~\ref{Subject-setup}(c)--(d). It consisted of a visual guidance interface, a pneumatic control box, and a laptop computer serving as the central controller.

The pneumatic control box included a small air reservoir (CRVZS-0.75, Festo), a custom ESP32-based printed circuit board (PCB), a 24~V lithium battery, and two proportional pressure regulators (ITV0030, SMC). The 0.75~L air tank buffered and stabilized input pressure, while the custom PCB controlled and monitored output pressure of proportional valves via analog signals. The system was supplied by a portable commercial air compressor (F6000/50, FIAC).

A MATLAB-based graphical user interface (GUI) was developed to provide visual and auditory cues, ensuring consistency across repetitions. The GUI displayed a simplified rotating bar to guide shoulder motion, accompanied by textual and auditory prompts indicating different motion phases. During experiments, once operating parameters were configured, the GUI simultaneously provided visual and auditory guidance to participants and transmitted pressure control commands to the pneumatic control box via serial communication, enabling real-time regulation of exoskeleton assistance.

\subsection{Data Acquisition and Processing}

During dynamic tests, seven sEMG sensors were placed on biceps brachii (BIC), supraspinatus (SS), upper trapezius (UT), clavicular head of pectoralis major (PM), anterior deltoid (AD), posterior deltoid (PD), and middle deltoid (MD). For BIC, SS, and UT, three Avanti sensors were used with 2048~Hz sampling frequency, while the other four muscles used a Quattro sensor with 1926~Hz sampling. Before experiments, maximum voluntary contraction (MVC) of muscles was recorded as the baseline for muscle activity evaluation.

After sEMG data acquisition, all signals were preprocessed following a standard pipeline, including digital filtering for noise reduction, full-wave rectification, smoothing using a moving-average filter with 0.2~s window, and normalization based on MVC measurements~\cite{Rui2025Glove,Rui2025LPPAMs}. Subsequently, feature extraction was performed. For each subject and experimental condition, sEMG signals were manually segmented to extract muscle activity corresponding to the arm-raising phase of each repetition, from onset of arm elevation to completion of arm lowering, with duration of approximately 13~s. Mean value of the segmented signal was computed for each repetition. Finally, results from five repetitions were averaged to obtain a single representative value per condition for subsequent statistical analysis.

After feature extraction, statistical analysis was conducted on data collected from ten subjects. Considering the relatively small sample size and observed outliers in data, non-parametric statistical tests were applied to all muscle activity results to ensure consistency across analyses. For shoulder abduction, where only two conditions were compared, two-tailed Wilcoxon signed-rank tests were used with significance denoted as **: $p < 0.01$; *: $p < 0.05$. For shoulder flexion, where three conditions were evaluated, Friedman tests were first applied, followed by post-hoc pairwise comparisons using two-tailed Wilcoxon signed-rank tests with Bonferroni correction ($\alpha_{\text{corrected}} = 0.0167$), with significance denoted as **: $p < 0.0167$ (significant); *: $0.0167 \leq p < 0.05$ (trend, not significant after correction).

Muscle activity reduction was quantified based on median values and calculated as $\text{Reduction} = 100\% \times (\text{No exo} - \text{Exo})/{\text{No exo}}$. Muscle activity reduction values were reported in figures only when statistically significant differences were observed.

For ROM analysis, Wilcoxon signed-rank tests were applied. Mean values were indicated in figures, and ROM reduction was calculated based on median values.

At the end of experiments, participants completed a questionnaire to evaluate their subjective experience and perception of the exosuit. The questionnaire included nine items:

\begin{enumerate}
\item How fatigued did you feel during flexion movements without any assistance? (1 = extremely fatigued, 7 = not fatigued)
\item How fatigued did you feel during flexion movements with only abduction assistance?
\item How fatigued did you feel during flexion movements with both types of assistance?
\item How fatigued did you feel during abduction movements without any assistance?
\item How fatigued did you feel during abduction movements with assistance?
\item How easy was it to put on and take off the exoskeleton? (1 = very difficult, 7 = very easy)
\item To what extent did the exoskeleton restrict your arm movements? (1 = very restricted, 7 = completely free)
\item How comfortable did you feel while wearing the exoskeleton? (1 = very uncomfortable, 7 = very comfortable)
\item How noticeable was the assistance provided by the exoskeleton? (1 = not noticeable at all, 7 = very noticeable)
\end{enumerate}

The first five items primarily assessed perceived fatigue across different exoskeleton conditions, while the last four items evaluated donning/doffing, movement restriction, comfort, and perceived assistance. Participants rated each item on a 7-point Likert scale, where higher scores indicated more favorable experience. For questionnaire data, Wilcoxon signed-rank tests were applied.

\subsection{Use Study Results}

\begin{figure*}[tbhp]
\centering
\includegraphics[width=1\linewidth]{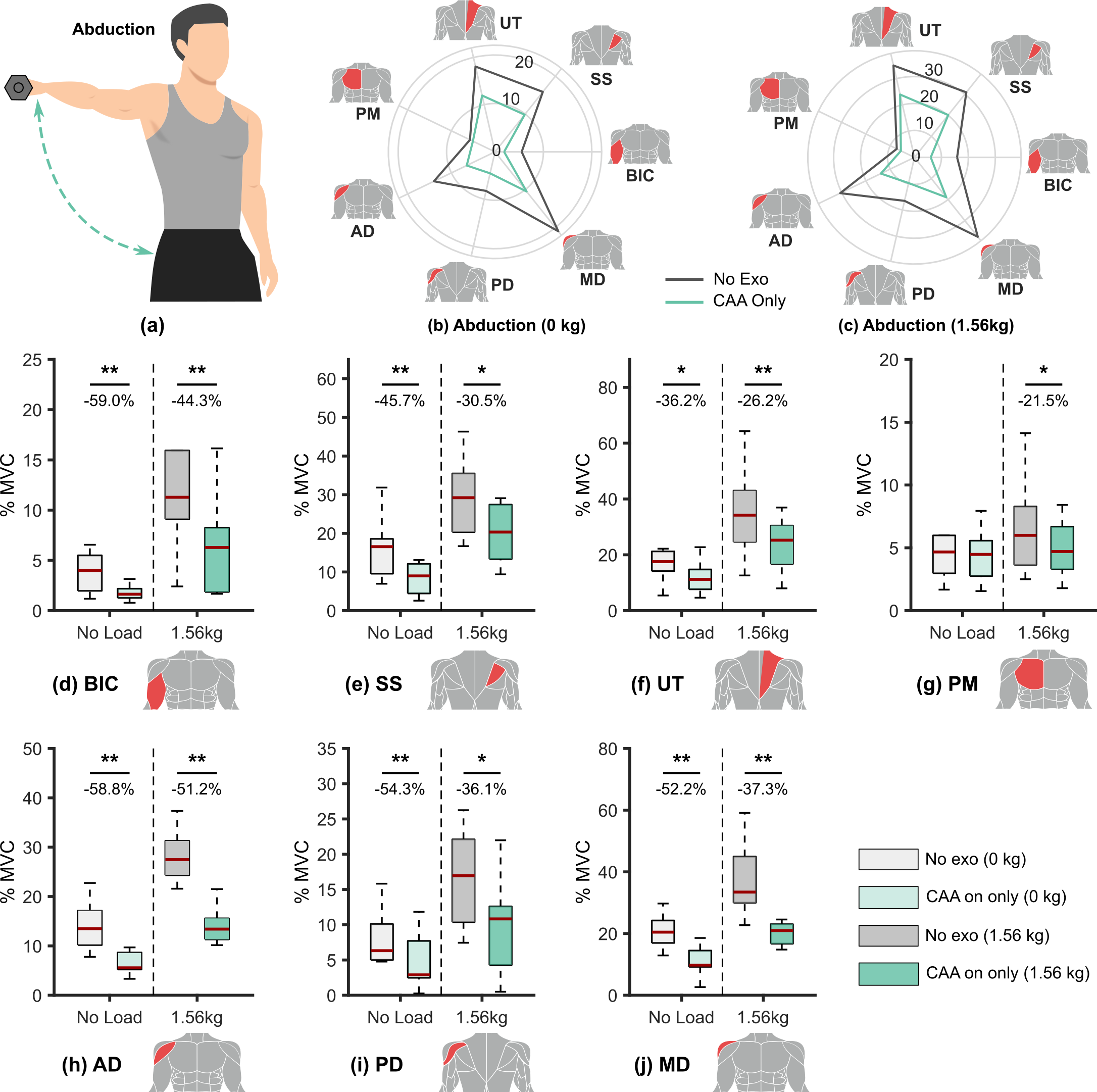}
\caption{Muscle activity comparison during shoulder abduction movements under different exosuit and weight conditions. (a) Abduction movement with weight. (b-c) Polar plot of muscle activation patterns during shoulder abduction without load and with load. (d-j) Statistical results for each muscle during shoulder abduction movements under different exosuit and weight conditions. Statistical analysis: Wilcoxon signed-rank tests (**: $p < 0.01$; *: $p < 0.05$).}
\label{Subject-Abduction}
\end{figure*}

For flexion, we compared three exoskeleton conditions (Fig.~\ref{Subject-flexion}a): no exosuit, CAA only, and CAA + HAA, representing no exosuit support, abduction actuator support, and both actuators support, respectively (raw sEMG data of one representative subject provided in Supplementary Material Figure 1). Polar plots of muscle activation patterns during shoulder flexion without load and with load are shown in Fig.~\ref{Subject-flexion}(b-c) (mean results). Compared to no exosuit, all muscles showed reduction in sEMG, while the two conditions with exosuit showed slightly different patterns. When both CAA and HAA were on without load, muscle activity of PM was smaller but UT was slightly greater compared to CAA only, while other muscles showed negligible differences. With load, both CAA and HAA on showed greater muscle activity reduction in PM and SS, while more AD was required.

\begin{figure*}[tbhp]
\centering
\includegraphics[width=1\linewidth]{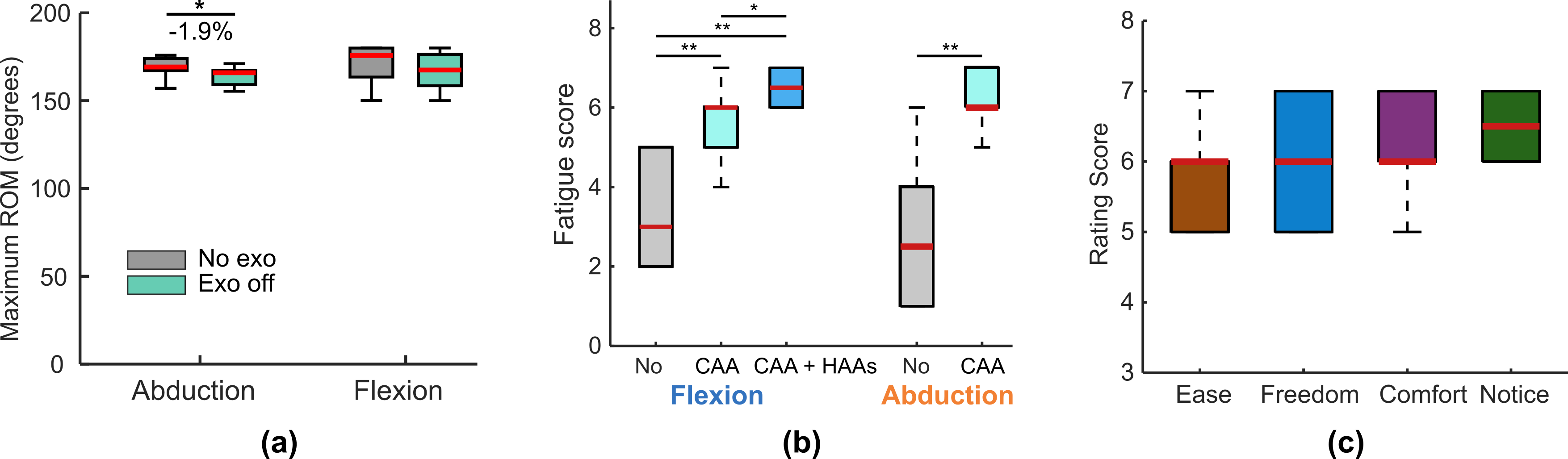}
\caption{ROM comparision and user evaluation results. (a) Comparison of ROM with and without the exosuit. (b) Fatigue scores for flexion and abduction movements. (c) User ratings for ease of donning/doffing, freedom of motion, comfort, and assistance effectiveness.}
\label{Questinaire}
\end{figure*}

Statistical analysis across seven muscles under different loads and exosuit conditions is shown in Fig.~\ref{Subject-flexion}(d-j) (detailed statistical values provided in Supplementary Material Tables 1-2). Both exosuit conditions significantly (after correction) reduced muscle activity for most muscles (24/28 comparisons). Compared to unloaded conditions, muscle activity reduction for BIC, UT, and MD was smaller under loaded conditions. For SS and PM, relative reductions were largely unchanged, while for AD and PD, reductions were greater under loaded conditions.

Specifically, for BIC under unloaded conditions, muscle activity was reduced by up to 57.4\% ($p = 0.0039$, $d = 0.89$) with exosuit support, whereas under loaded conditions maximum reduction was 39.7\% ($p = 0.0098$, $d = 0.77$). Similarly, maximum reductions for UT, PM, and MD decreased from 54.5\% ($p = 0.0195$, $d = 0.75$) to 28.5\% ($p = 0.0059$, $d = 0.84$), 29.9\% ($p = 0.002$, $d = 0.62$) to 29.2\% ($p = 0.002$, $d = 0.74$), and 62.0\% ($p = 0.002$, $d = 1.27$) to 44.7\% ($p = 0.002$, $d = 1.08$), respectively. SS, AD, and PD showed the opposite trend, with reductions increasing from 63.2\% ($p = 0.002$, $d = 0.77$), 26.8\% ($p = 0.0039$, $d = 0.8$), and 35.1\% ($p = 0.002$, $d = 0.51$) to 63.7\% ($p = 0.0039$, $d = 1$), 35.3\% ($p = 0.002$, $d = 0.61$), and 41.9\% ($p = 0.002$, $d = 0.95$), respectively.

For both loading conditions, differences between abduction-only and dual-actuator conditions were not pronounced. Significant differences were observed only in a few muscles. Although data suggest that for BIC, AD, and PD, dual-actuator activation could reduce more muscle activity than single-actuator, statistical significance was not reached. The difference between two actuators and one actuator was significant only for PM, showing reductions of --14.1\% ($p = 0.002$, $d = 0.46$) and --13.4\% ($p = 0.0371$, $d = 0.28$) under unloaded and loaded conditions, respectively. For SS and UT, dual-actuator compared to single-actuator increased muscle activity by 48.2\% ($p = 0.0039$, $d = 0.8$) and 35.6\% ($p = 0.0488$, $d = -0.22$), respectively. 

For abduction, statistical analysis of seven muscles under different loading and exosuit conditions is presented in Fig.~\ref{Subject-Abduction} (raw sEMG data of one representative subject provided in Supplementary Material Figure 1; detailed statistical values in Supplementary Material Tables 1-3). Except for PM under unloaded conditions where no significant difference was observed, all other muscles showed significant muscle activity reduction with exosuit support, regardless of loading condition.

Under unloaded conditions, BIC exhibited the largest reduction, reaching 59\% ($p = 0.002$, $d = 0.8$). Primary muscles contributing to abduction, including AD, PD, and MD, all showed reductions exceeding 50\%. For SS, reduction was 45.7\% ($p = 0.002$, $d = 0.81$), while UT, which exhibited the smallest relative change, still showed substantial reduction of 36.2\% ($p = 0.0371$, $d = 0.9$).

Under loaded conditions, compared to unloaded cases, magnitude of muscle activity reduction decreased for all muscles. AD showed the largest reduction at 51.2\% ($p = 0.002$, $d = 1.52$), corresponding to more than 14\% MVC decrease in muscle activity. For BIC, MD, and PD, relative reductions also exceeded 35\%, with values of 44.3\% ($p = 0.002$, $d = 0.95$), 37.3\% ($p = 0.002$, $d = 1.99$), and 36.1\% ($p = 0.0137$, $d = 0.96$), respectively. PM exhibited the smallest muscle activity reduction at approximately 21.5\% ($p = 0.0137$, $d = 0.42$). Overall, these results demonstrate that the exosuit provides effective support for abduction movements and substantially reduces required muscle activity.

Experimental results for ROM are presented in Fig.~\ref{Questinaire}(a). For abduction, median maximum ROM without exoskeleton was 169.18$^\circ$, while with exoskeleton it was 165.93$^\circ$, representing 1.9\% relative reduction. Although statistically significant ROM reduction was observed ($p = 0.027$, $d = 0.767$), the magnitude was negligible. For flexion, no significant ROM reduction was found.

Questionnaire results are shown in Fig.~\ref{Questinaire}(b-c). For flexion, significance was observed between three conditions (No vs Abd: $p = 0.002$, $d = -2.424$; No vs Full: $p = 0.002$, $d = -3.54$; Abd vs Full: $p = 0.015$, $d = -1.265$). With both actuators on, subjects felt less fatigue than with abduction actuator only, and more fatigue without exosuit support. For abduction, significant fatigue reduction demonstrated assistance effectiveness of the abduction actuator and exosuit ($p = 0.002$, $d = -1.255$). All subjects rated exosuit performance highly, with median scores of 6, 6, 6, and 6.5 for ease of donning/doffing, freedom of movement, comfort, and shoulder assistance, respectively.

\section{Discussion}

\subsection{Spindle-Shaped Actuator Design Methodology}

The SSAA design is motivated by a fundamental observation: in pneumatic bending actuators, dynamic response is governed primarily by the volume of air required for inflation, rather than by the actuator's output force. By tapering the cross-sections toward both ends while keeping the fold section unchanged, the SSAA reduces inflation volume without sacrificing peak force output. This trade-off is well-suited to shoulder abduction, where torque requirements vary nonlinearly across the range of motion. The close agreement between volume reduction and response time improvement confirms that the performance gain comes primarily from the reduced air volume rather than from changes in material compliance or bending kinematics.

In this work, $w_2$ was held constant while $w_1$ and $w_3$ were reduced to isolate the effect of volume reduction on dynamic response, serving as a controlled comparison against the uniform cylindrical baseline. Within this design framework, $w_2$, $w_1$, and $w_3$ represent partially independent degrees of freedom: increasing $w_2$ strengthens moment output, while reducing $w_1$ and $w_3$ preserves the volumetric advantage. Exploring this joint optimization space is a natural direction for future work, particularly for applications where both peak torque and bandwidth are constrained.

One boundary condition deserves attention: the distal end must remain large enough to distribute contact forces to the axillary region without buckling under load. At higher abduction angles, where the actuator approaches full inflation, contact forces at the ends increase, and an overly tapered geometry may collapse under these loads. The minimum viable end dimension is therefore application-specific and should be validated for the target user population prior to deployment.

The spindle-shaped design is not limited to shoulder applications. Any fabric-based bending actuator where response speed matters and torque requirements vary across the range of motion could benefit from this approach.

\subsection{System Integration and Characterization}

The Y-shaped CAA design offers two functional advantages over a single straight actuator. First, during abduction, the Y-bifurcation provides more stable contact with the arm: a single actuator with insufficient constraint may slide laterally under load, compromising consistent force transmission. The three additional constraint layers in the CAA prevent this lateral shift and maintain reliable contact throughout the motion. Second, during horizontal adduction, the Y-shape distributes the adduction force more effectively across the attachment area, enabling better moment transmission compared to a single-axis actuator. The CAA achieves a maximum moment of 9.7~Nm at $\theta_a = 90^{\circ}$ and 90~kPa, slightly higher than a comparable shoulder exosuit actuator\footnote{Peak moment estimated from figures reported in the compared work.} (8.3~Nm)~\cite{Campioni2025_Preliminary_clinical_Evaluation}. With two SSAA units, the total actuator volume is 714~mL, which is 28.6\% smaller than the compared work (1~L)~\cite{Campioni2025_Preliminary_clinical_Evaluation}, enabling faster response for a given air supply or a lighter portable system.

The moment-angle characteristic of the CAA shows relatively stable output for the single SSAA, but declining output at low abduction angles for the combined actuator, which is consistent with geometric overlap between the two branches at low bending angles. This is not a meaningful functional limitation, however, since gravitational loading on the arm decreases proportionally at low abduction angles, and the reduction in actuator output in this range does not significantly compromise gravitational compensation. Considering the anchoring position of the CAA, the torque applied to the shoulder joint is expected to exceed the actuator's measured output. Based on the method outlined in~\cite{proietti2023_shoulder_exosuit_restoring_Science, zhou2024portable_industry}, the required torque for raising the arm depends on the user's height and weight~\cite{plagenhoef1983anatomical1, fryar2012anthropometric2, dempster1955anthropometric3}. Nevertheless, the 9.7~Nm peak output of the CAA can provide considerable gravitational support for the majority of users across the functional range of shoulder abduction.

For the HAA, the two-stage modeling approach---first characterizing individual pouch motor force--height behavior, then mapping to the moment--angle space of the assembled HAA---introduces some smoothing error, since real actuators do not exhibit the discrete transition assumed in the model. In addition, actuator--actuator and actuator--body interactions were neglected, which may lead to overestimation of the achievable bending angle $\beta$ in practice.

\subsection{Functional Characterization in Healthy Users}

The user study results confirm that the volume-optimized actuators provide meaningful gravitational compensation during dynamic upper-limb tasks. The EMG reductions observed during abduction are consistent with effective torque assistance, and their persistence under loaded conditions suggests the exosuit remains effective as task demand increases.

The more informative finding comes from comparing single- and dual-actuator conditions during flexion. The limited additional benefit of the HAA in healthy participants is most likely explained by the intact neuromuscular control of this population. Healthy users can rapidly adapt their motor patterns in response to novel mechanical inputs, and the additional forces introduced by the HAA may trigger compensatory co-contraction to preserve joint stability rather than reducing muscle effort. The increased activity in SS and UT under dual-actuator conditions supports this interpretation: these muscles stabilize the scapula, and their elevated recruitment likely reflects active regulation of glenohumeral geometry in response to the unfamiliar force environment.

This finding has a direct implication for target population selection. In individuals with neurological motor impairments, where voluntary compensatory responses are reduced, the same assistance may produce substantially different outcomes. Weakened muscles may respond more directly to external mechanical support, reducing the likelihood of compensatory co-contraction and potentially making multi-DOF assistance more beneficial. Whether this holds empirically remains the most important validation step before clinical translation.

From a control perspective, the results suggest that multi-DOF assistance may be more effective with task-aware activation strategies rather than continuous simultaneous actuation of all degrees of freedom. Selectively activating the HAA during the horizontal adduction phase---rather than throughout the entire flexion movement---may capture the mechanical benefit while reducing the window during which stabilizer co-contraction is elicited.

\subsection{Design Implications and Limitations}

The consistent effectiveness of the abduction actuator across participants with varied anthropometry suggests that the Y-shaped CAA is sufficiently robust for general use without per-subject customization, which is practically important for both clinical and industrial deployment. The limited incremental benefit of adding the HAA in healthy users suggests that for routine gravitational offloading tasks, a single-DOF design may achieve most of the functional goal at lower system complexity, weight, and cost.

Several limitations should be noted. The geometric optimization of the SSAA remains partially unexplored: systematic parametric sweeps over end and fold dimensions would allow the trade-off between volume, moment, and response time to be characterized more completely. The mathematical models for both actuators exclude inter-actuator and actuator--body coupling, limiting their predictive accuracy in configurations where these interactions are significant. Durability beyond 2000 cycles has not been assessed, and long-term fatigue behavior of the fabric and seam construction will be critical for clinical use. Most importantly, all functional data were collected from healthy participants; validation in populations with motor impairments, as well as testing under isometric conditions and horizontal adduction tasks, is needed before the findings can be extended to clinical contexts.

\section{Conclusion}

This work presents a design methodology for dual-DOF soft shoulder exosuits addressing portable pneumatic system constraints. The spindle-shaped angled actuator (SSAA) addresses the torque--response trade-off inherent in portable pneumatic systems: by reducing actuator volume by 35.7\%, it achieves faster dynamic response while maintaining 94.2\% of torque output, enabling more efficient use of the limited air supply resources critical for portable operation. Building on the SSAA, a curved abduction actuator (CAA) and a pouch-motor-based horizontal adduction actuator (HAA) were developed and integrated into a dual-DOF textile-based shoulder exosuit (390~g), capable of delivering multi-modal assistance across shoulder abduction, flexion, and horizontal adduction.

User studies with 10 healthy participants demonstrated substantial muscle activity reductions across both abduction and flexion tasks. For abduction, the exosuit achieved up to 59\% EMG reduction across seven muscles, validating effective gravitational compensation. For flexion, both the single-actuator configuration (HAA only) and the dual-actuator configuration (HAA\,+\,CAA) reduced EMG activity by up to 63.7\% compared to no assistance. However, the incremental benefit of adding the CAA to existing HAA support was limited in healthy users during flexion, with statistically significant additional reductions observed only in pectoralis major. Validation in populations with impaired neuromuscular control is needed to determine whether dual-actuator assistance provides greater incremental benefit in clinical applications.

These empirical findings characterize actuator contributions in healthy users and provide quantitative data to inform design decisions for multi-DOF exosuit systems. The volume-optimized actuator design methodology applies beyond shoulder assistance to other pneumatic bending actuators where moment arm analysis reveals non-uniform force requirements, such as elbow exosuits, soft grippers, and continuum manipulators. Future work should include validation in relevant user populations and investigation of task-specific control strategies for multi-DOF assistance.

\section*{Acknowledgments}

We sincerely thank all participants for their time and contribution to this study.

\bibliographystyle{IEEEtran}
\bibliography{mybib}

\vfill

\end{document}